# Data augmentation approaches for improving animal audio classification

Loris Nanni[a] Gianluca Maguolo[a*] Michelangelo Paci[b]
[a] DEI, University of Padua, viale Gradenigo 6, Padua, Italy. loris.nanni@unipd.it
[b] BioMediTech, Faculty of Medicine and Health Technology, Tampere University, Arvo Ylpön katu 34, D 219, FI-33520, Tampere, Finland
[*] Corresponding author, email: gianluca.maguolo@phd.unipd.it

**Abstract.** In this paper we present ensembles of classifiers for automated animal audio classification, exploiting different data augmentation techniques for training Convolutional Neural Networks (CNNs). The specific animal audio classification problems are i) birds and ii) cat sounds, whose datasets are freely available. We train five different CNNs on the original datasets and on their versions augmented by four augmentation protocols, working on the raw audio signals or their representations as spectrograms. We compared our best approaches with the state of the art, showing that we obtain the best recognition rate on the same datasets, without ad hoc parameter optimization. Our study shows that different CNNs can be trained for the purpose of animal audio classification and that their fusion works better than the stand-alone classifiers. To the best of our knowledge this is the largest study on data augmentation for CNNs in animal audio classification audio datasets using the same set of classifiers and parameters. Our MATLAB code is available at https://github.com/LorisNanni.

## 1. Introduction

In the current context of constantly increasing environmental awareness, highly accurate sound recognition systems can play a pivotal role in mitigating or managing threats like the increasing risk of animal species loss or climate changes affecting the wildlife fauna [1]. Sound classification and recognition has been included among the pattern recognition tasks for different application domains, e.g. speech recognition [2], music classification [3], environmental sound recognition or biometric identification [4]. In the traditional pattern recognition framework (preprocessing, feature extraction and classification) features have generally been extracted from the actual audio traces (e.g. Statistical

Spectrum Descriptor or Rhythm Histogram [5]). However, the conversion of audio traces into their visual representations enabled the use of feature extraction techniques commonly used for image classification. The most common visual representation of audio traces displays the spectrum of frequencies of the original traces as it varies with time, e.g. spectrograms [6], Mel-frequency Cepstral Coefficients spectrograms [7] and other representations derived from these. A spectrogram can be described as a bidimensional graph with two geometric dimensions (time and frequency) plus a third dimension encoding the signal amplitude in a specific frequency at a particular time step as pixel intensity [8]. For example, Costa et al. [9,10] applied many texture analysis and classification techniques to music genre classification. In [10] the grey level co-occurrence matrices (GLCMs) [11] were computed on spectrograms as features to train support vector machines (SVMs) on the Latin Music Database (LMD) [12]. Similarly, in [9] they used one of the most famous texture descriptor, the local binary pattern (LBP) [13], again to train SVMs on the LMD and the ISMIR04 [14] datasets, improving the accuracy of their classification with respect to their previous work. Again in 2013 [15], they used the same approach, but using local phase quantization (LPQ) and Gabor filters [16] for feature extraction. This actually marked an interesting parallel in the development of more and more refined texture descriptors for image classification and their application also to sound recognition. In 2017, Nanni et al. [3] presented the fusion of state-of-the-art texture descriptors with acoustic features extracted from the audio traces on multiple dataset, demonstrating how such fusion greatly improved the accuracy of a system based only on acoustic or visual features. However, with the diffusion of deep learning and the availability of more and more powerful Graphic Processing Units (GPUs) at accessible costs, i) the canonical pattern recognition framework changed and ii) the attention was polarized on visual representations of acoustic traces. The optimization of the feature extraction step had a key role in the canonical framework, especially with the development of handcrafted features that place patterns from the same class closer to each other in the feature space, simultaneously maximizing their distance from other classes. Since deep classifiers learn the best features for describing patterns during the training process, the aforementioned feature engineering lost part of its importance and it has been coupled with the direct use of the visual representation of audio traces, letting the classifiers selecting the most informative features. Another reason for representing the patterns as images at the beginning of the pipeline is the intrinsic architecture of the most famous deep classifiers, such as convolutional neural networks (CNN), which require images as their input. This motivated researchers using CNNs in audio classification to advance methods for the conversion of audio signals into time-frequency images.

Among the first studies using deep learning for audio images, Humphrey and Bello [17,18] explored CNNs as alternatives to addressed music classification problems, defining the state of the art in automatic chord detection and recognition. Nakashika et al. [19] performed music genre classification on the GTZAN dataset [20] converting spectrograms into GCLM maps to train CNNs. Costa et al. [21] fused canonical approaches, e.g. LBP-trained SVMs with CNNs, performing better that the state of the art on the LMD dataset.

In addition to approaches derived directly from image classification, few studies focused on different classification aspects, in order to make such process more specific for sound recognition. Sigtia and Dixon [22] aimed to adjust CNN parameters and structures, and showed how the training time was reduced by replacing sigmoid units with Rectified Linear Units (ReLu) and stochastic gradient descent with the Hessian Free optimization. Wang et al. [23] proposed a novel CNN called a sparse coding CNN for sound event recognition and retrieval, obtaining competitive and sometimes better results than most of the other approaches when evaluating the performance under noisy and clean conditions. Another hybrid approach by Oramas et al. [24] combined different modalities (album cover images, reviews and audio tracks) for multi-label music genre classification using deep learning methods appropriate for each modality and outperforming the unimodal methods.

The clear improvement in classification performances introduced by the use of deep classifiers, led to apply sound recognition also to other tasks, such as the biodiversity assessment or monitoring animal species at risk. . For example, birds have been acknowledged as biological indicators for ecological research. Therefore, their observation and monitoring are increasingly important for biodiversity conservation, with the additional advantage that the acquisition of video and audio information is minimally invasive. To date, many datasets are available to develop classifiers to identify and monitor different species such as birds [25,26], whales [27], frogs [25], bats [26], cats [28]. For instance, Cao et al. [29] combined a CNN with handcrafted features to classify marine animals [30] (the Fish and MBARI benthic animal dataset [31]). Salamon et al. [32] investigated the use of fusing deep learning (using CNN) and shallow learning for the problem of bird species identification, based on 5,428 bird flight calls from 43 species. In both these works, the fusion of CNNs with mode canonical techniques outperformed the single approach.

One of the main drawbacks of deep learning approaches is the need of great amount of training data [33], in this case audio signals and consequently their visual representations. In case of limited amount of training images, data augmentation is a powerful tool. Animal sound datasets are usually much smaller than necessary, since the sample collection and labelling can be very expensive. Commonly, audio signals can be augmented in the time and/or in the frequency domains directly on the raw signals or after their conversion into spectrograms. In [34] different augmentation techniques were applied to the training set for the BirdCLEF 2018 initiative (www.imageclef.org/node/230) that included over 30,000 bird sound samples ranging over 1,500 species. Bird audio signals were first augmented in the time domain by e.g. extracting chunks from random position in each file, applying jitter to duration, add two audio chunks from random files background noise and background atmospheric noise, applying random cyclic shift and time interval dropout. Every augmented audio chunk was then converted into spectrogram and then further augmented in the frequency domain by pitch shift and frequency stretch, piecewise time stretch and frequency stretch and applying color jittering. The influence of the complete augmentation led improve by almost 10% the identification performance quantified as Mean Reciprocal Rank. In the field of animal audio classification, Sprengel et al. [35] used standard audio augmentation techniques for bird audio classification, such as time and pitch shift. Besides, they created more samples by summing two different samples belonging to the same class. This is motivated by the fact that the sound of two birds from the same class should still be correctly classified. Pandeya et al. [28] demonstrated that audio signal augmentation by simple techniques as random selection of time stretching, pitch shifting, dynamic range compression, and insertion of noise on the domestic cat sound dataset, described in Section 5 of this paper, improved accuracy, F1-score and area under ROC curve. In particular, the performance improvement increased by including more augmented clones (one to three) per single original audio file. Conversely, Oikarinen et al. [36], showed that augmenting their spectrograms by translations, adding random noise, and multiplying the input by a random value close to one, did not significantly improve their classification of marmoset audio signals. Of note, the aim of Oikarinen et al. was not the classification of species or call types only, e.g. from publicly available datasets, but the identification of call types and the source animal in a complex experimental setup consisting of multiple cages in one room, each cage containing two marmosets. Other techniques, inherited from e.g. speech recognition, are also suitable for animal sound classification. For instance, Jaitly et al. [37] proposed Vocal Track Length Perturbation (VTLP), which alters the vocal tract length during the extraction of a descriptor to create a new sample. They show that this technique is very effective in speech recognition. Takahashi et al. [38] used large convolutional networks with strong data augmentation to classify audio events. They also used VTLP and introduced a new transformation that consists in summing two different perturbed samples of the same class.

In this work, we compare different sets of data augmentation approaches, each coupled with different CNNs. This way, an ensemble of networks is trained. Finally, the set of classifiers is combined by sum rule. The proposed method is tested in two different audio classification dataset: the first related to domestic cat sound classification ([28]), the latter on bird classification ([1]). Our experiments were designed to compare and maximize the performance obtained by varying combinations of data augmentation approaches and classifiers and they showed that our augmentation techniques were successful at improving the classification accuracy.

Our main contributions to the community are the following:

- Different methods for audio data augmentation are tested/proposed/compared in two datasets;
- Exhaustive tests are performed on the fusions among ensemble system based on CNNs trained with different data augmentation approaches;
- All MATLAB source code used in our experiments will be freely available at https://github.com/LorisNanni

## 2. Audio Image Representation

In order to get image representations for the audio signals we applied a Discrete Gabor Transform (DGT) to the signal. The DGT is a particular case of Short-Time Fourier Transform where the window function is a Gaussian kernel. The continuous Gabor transform is defined as the convolution between a Gaussian and the product of the signal with a complex exponential:

$$G(\tau,\omega) = \frac{1}{\sigma^2}\int_{-\infty}^{+\infty} x(t)e^{i\omega t}e^{-\pi\sigma^2(t-\tau)^2}\ dt$$

where $x(t)$ is the signal, $\omega$ is a frequency and $i$ is the imaginary unit. The parameter $\sigma^2$ is the width of the Gaussian window. The discrete version of the DGT uses the discrete convolution. The output $G(\tau,\omega)$ is a matrix whose columns represent the frequencies of the signal at a fixed time. We used the DGT implementation provided in http://ltfat.github.io/doc/gabor/sgram.html [39].

## 3. Convolutional Neural Networks

In this work, we used CNNs both for feature extraction (to train SVMs) and for direct classification. CNNs, introduced in 1998 by LeCun et al. [40], are deep feed-forward neural networks where neurons are connected only locally to neurons from the previous layer. Weights, biases and activation functions are iteratively adjusted during the training phase. In addition to the input layer, i.e. the image or its part to be classified, and the output/classification (CLASS) layer, composed by one neuron for each class to classify, a CNN contains one or more hidden layers. The different types of hidden layers are convolutional (CONV), activation (ACT), pooling (POOL) and fully-connected (FC). The CONV layers perform feature extraction from the input volume by convolving a local region of the input volume (receptive field) to filters of the same size, thus a single integer of the output volume (feature map). Then the filter slides over the next receptive field of the same input image by a defined stride and again the convolution between the new receptive field and the same filter is computed. Doing this for the whole input image provides the input for the next layer. After each CONV layer, a non-linear ACT layer is applied to improve classification and the learning capabilities of the network. Common activation functions are the non-saturating ReLU function $f(x) = \max(0, x)$ or the saturating hyperbolic tangent $f(x) = \tanh(x)$, $f(x) = |\tanh(x)|$, or the sigmoid function $f(x) = (1 + e^{-x})^{-1}$. POOL layers are required to perform non-linear downsampling operations (e.g. max or average pool) aimed at reducing the spatial size of the representation while simultaneously decreasing 1) the number of parameters, 2) the possibility of overfitting, and 3) the computational complexity of the network. POOL layers are commonly present between two CONV layers. FC layers are usually the last hidden layers: they have neurons fully connected to all the activations in the previous layer. The output CLASS layer performs the final classification: SoftMax is a commonly used activation function for the CLASS layer.

We adapt CNNs that were previously pre-trained on ImageNet [41] or Places365 [42] datasets to our classification problems. In detail, we keep the original pre-trained network architectures but the last three layers are replaced by i) an FC layer, ii) an ACT layer using SoftMax and iii) a CLASS layer. Then we retrain the whole CNNs, using as starting values for the network weights, the original values of the pre-trained networks. We test and combine two different CNN architectures:

1. GoogleNet [43]. This CNN is the winner of the ImageNet Large Scale Visual Recognition

Competition (ILSVRC) challenge in 2014. Its structure includes 22 layers that require training and five POOL layers. It also introduces a new "Inception" module (INC), i.e. a subnetwork made of parallel convolutional filters whose outputs are concatenated, greatly reducing the amount of learnable parameters. Two pre-trained GoogleNets are used: the one trained on ImageNet database [41] and the one trained on the Places365 [42] datasets.

2. VGGNet [44]. This CNN placed second in ILSVRC 2014. It is a very deep network that includes 16 (VGG-16) or 19 (VGG-19) CONV/FC layers. The CONV layers are extremely homogeneous and use very small (3x3) convolutional filters with a POOL layer after every two or three CONV layers (instead after each CONV layer as in e.g. AlexNet [45]). Both VGG-16 and VGG-19 are trained on ImageNet database [41].

## 4. Data Augmentation approaches

In this paper, we tested the following four augmentation protocols. For the third and fourth protocols we used the methods provided Audiogmenter [46], an audio data augmentation library for MATLAB.

### 4.1 Standard Image Augmentation

Our first data augmentation protocol (StandardIMG, Fig. 2) combines standard data augmentation techniques in computer vision. We independently reflect the image in both the left-right (*RandXReflection*) and the top-bottom (*RandYReflection*) directions with 50% probability. We also linearly scale the image along both axes by two random numbers in [1, 2] (*RandXScale* and *RandYScale*). Besides, we apply random rotation by an angle in [-10, 10] (*RandRotation*) and a translation by a number of pixels in [0, 5] (*RandXTranslation* and *RandYTranslation*).

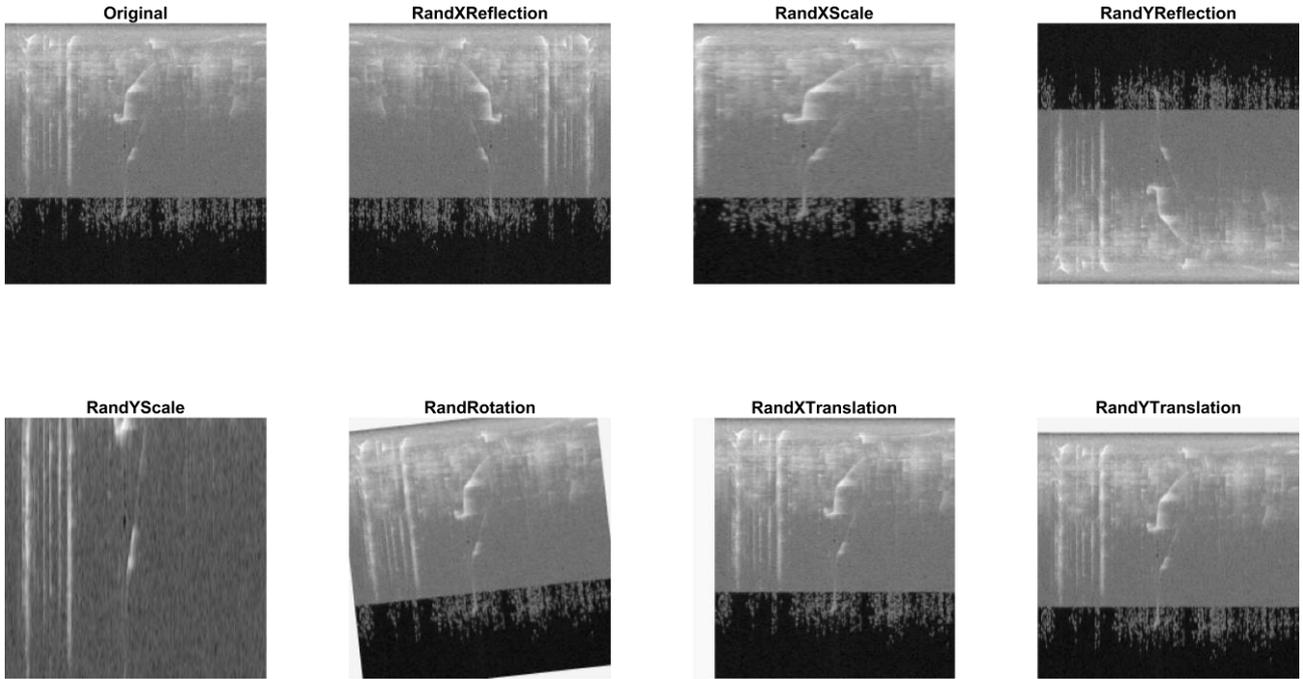

Figure 1. Effect of the seven standard image augmentation built-in in MATLAB techniques on one illustrative spectrogram produced from the original audio signal. To make more clear each transformation, to produce this figure the parameters were increased compared to those listed in Section 4.1 (scaling in [1, 10] and translation in [0, 20]).

**4.2 Standard Signal Augmentation**

Our second data augmentation protocol (StandardSGN) relies on the MATLAB built-in data augmentation methods for audio signals. We create 10 new signals for each training signal by applying the following transformations with 50% probability:

1. Signal speed scaling by a random number in [0.8, 1.2] (*SpeedupFactoryRange*).
2. Pitch shift by a random number in [−2,2] semitones (*SemitoneShiftRange*).
3. Volume increase/decrease by a random number in [−3,3] dB (*VolumeGainRange*).
4. Addition of random noise in the range [0, 10] dB (*SNR*).
5. Time shift in the range [−0.005, 0.005] seconds (*TimeShiftRange*).

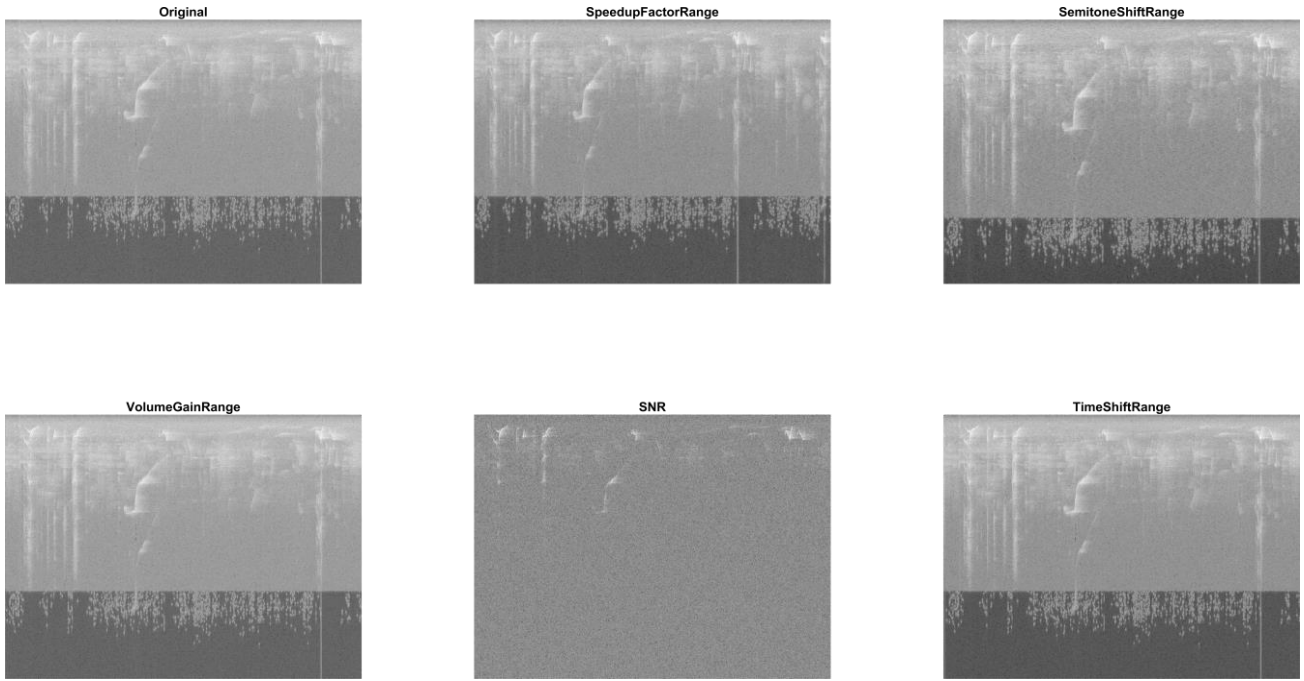

Figure 2. Effect of the five standard audio transformations built-in in MATLAB. Spectrograms were produced from the transformed audio signals.

### 4.3 Spectrogram Augmentation

Our third data augmentation protocol (Spectro, Fig. 3) works directly on spectrograms, producing six transformed versions of each original spectrogram. We implemented following six different functions (reported in italic):

1. *spectrogramRandomShifts* randomly applies pitch shift and time shift.

2. *spectrogramSameClassSum* creates a new image by summing the spectrograms of two random images from the same class.

3. Vocal Tract Length Normalization (*VTLN*) creates a new image by applying a random crop followed by a VTLP [37]. VTLP cuts the spectrogram into 10 different temporal slices and to each of them applies the formula

$$G(f) = \begin{cases} \alpha f, & 0 \leq f < f_0 \\ \dfrac{f_{max} - \alpha f_0}{f_{max} - f_0}(f - f_0) + \alpha f_0, & f_0 \leq f \leq f_{max} \end{cases}$$

where $f_0$, $f_{max}$ are the basic and maximum frequency, and $\alpha \in [a, b]$ is randomly chosen. We set *a* and *b* to 0.9 and 1.1, respectively.

4. *spectrogramEMDAaugmenter* applies the Equalized Mixture Data Augmentation (EMDA) [47] to create $n$ new images, where $n$ is the size of the original dataset, by computing the weighted average of two randomly chosen spectrograms with same label. We also apply i) a time delay, randomly selected in [0, 50], to one spectrogram and ii) a perturbation to both of them according to the formula $s_{aug}(t) = \alpha \Phi(s_1(t), \psi_1) + (1 - \alpha)\Phi(s_2(t - \beta T), \psi_2)$ where $\alpha, \beta$ are two random values in [0,1], $T$ is the time shift and $\Phi$ is an equalizer function parametrized by the vector $\psi = (f_0, g, Q)$. $f_0$ is the central frequency and it is randomly sampled in $[f_{0min}, f_{0max}] = [100, 6000]$. $g$ is the gain, which is randomly sampled in $[-GainMin, GainMax] = [-8, 8]$. The $Q$-factor $Q$ is randomly sampled in $[Qmin, Qmax] = [1, 9]$. All these parameters can be chosen by the user, the value here reported are those used in our experiments.

5. *randTimeShift* applies time shift by randomly picking the shift $T$ in $[1, M]$, where $M$ is the horizontal size of the input spectrogram, and cutting the spectrogram into two different images $S_1$ and $S_2$, taken before and after the time $T$. We obtain the new image by inverting the order of $S_1$ and $S_2$.

6. *randomImageWarp* applies Thin-Spline Image Warping [48] (TPS-Warp) to the spectrogram. TPS-Warp perturbs the original image by randomly changing the position of a subset $S$ of the input pixels and adapts the ones that do not belong to $S$ using a linear interpolation. We only change the spectrogram on the horizontal axis. Besides, we apply frequency and time masking, which is performed in practice by setting to zeros the entries of two rows and one column of the spectrogram. We set the width of the rows to 5 pixels and the width of the column to 15 pixels.

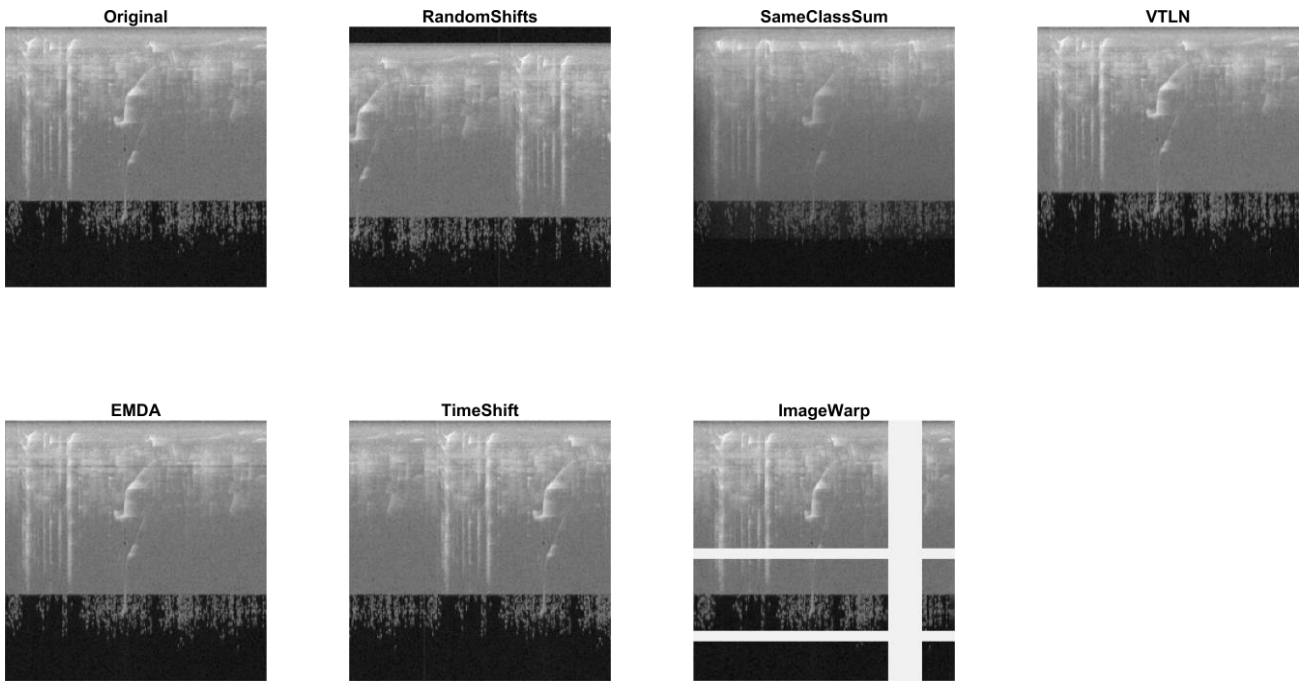

Figure 3. Effect of the six transformations included in the Spectro augmentation protocol and applied to one illustrative spectrogram after the raw audio signal conversion into visual representation.

**4.4 Signal Augmentation**

Our fourth protocol (Signal, Fig. 4) works directly on the raw audio signals, producing 11 transformed versions of the input signal. It consists in the following 10 functions (reported in italic):

1. *WowResampling* applies wow resampling to the original signal. Wow resampling is a variant of pitch shift where the intensity changes along time. The transformation is given by:

$$F(x) = x + a_m \frac{\sin(2\pi f_m x)}{2\pi f_m}$$

where x is the input signal, and we chose $a_m = 3$ and $f_m = 2$.

2. *Noise* adds white noise such that the ratio between the signal and the noise is $X$ dB, where $X$ can be chosen by the user. We used $X = 10$.

3. *Clipping* normalizes the audio signal leaving the 10% of the samples out of [-1, 1]. The out-of-range samples x are then clipped to sign(x).

4. *SpeedUp* increases or decreases the speed of the audio signal. In our experiments we applied a 15% speed augmentation.

5. *HarmonicDistortion* applies quadratic distortion to the signal 5 times consecutively:

$$s_{out} = \sin^5(2\pi s_{in})$$

where $\sin^5()$ represents the sine function applied five times.

6. *Gain* increases the gain of the audio signal by a specific number of dB. In our experiments we applied a 10 dB augmentation.

7. *randTimeShift* randomly breaks each audio signal in two parts, swaps them and mounts them back into a new randomly shifted signal, i.e. if $s_{in} = [s_0, s_x] \cup [s_x, s_{out}]$, the output signal is $s_{out} = [s_x, s_{out}] \cup [s_0, s_x]$.

8. *soundMix* sums two different audio signals from the same class to create a new synthetic signal.

9. *applyDynamicRangeCompressor* applies the Dynamic range compression (DRC) [49] to the input audio signal. DRC is a technique that boosts the lower intensities of an audio signal and attenuates the higher intensities according to an increasing and piecewise linear function, thus compressing the audio signal's dynamic range.

10. *pitchShift* shifts the pitch of an audio signal by a specific number of semitones. We chose to increase it and decrease it by two semitones. Fig. 4 reports two examples of pitch shift: pitchShiftA increases pitch by two semitones and pitchShiftB decreases it by two semitones.

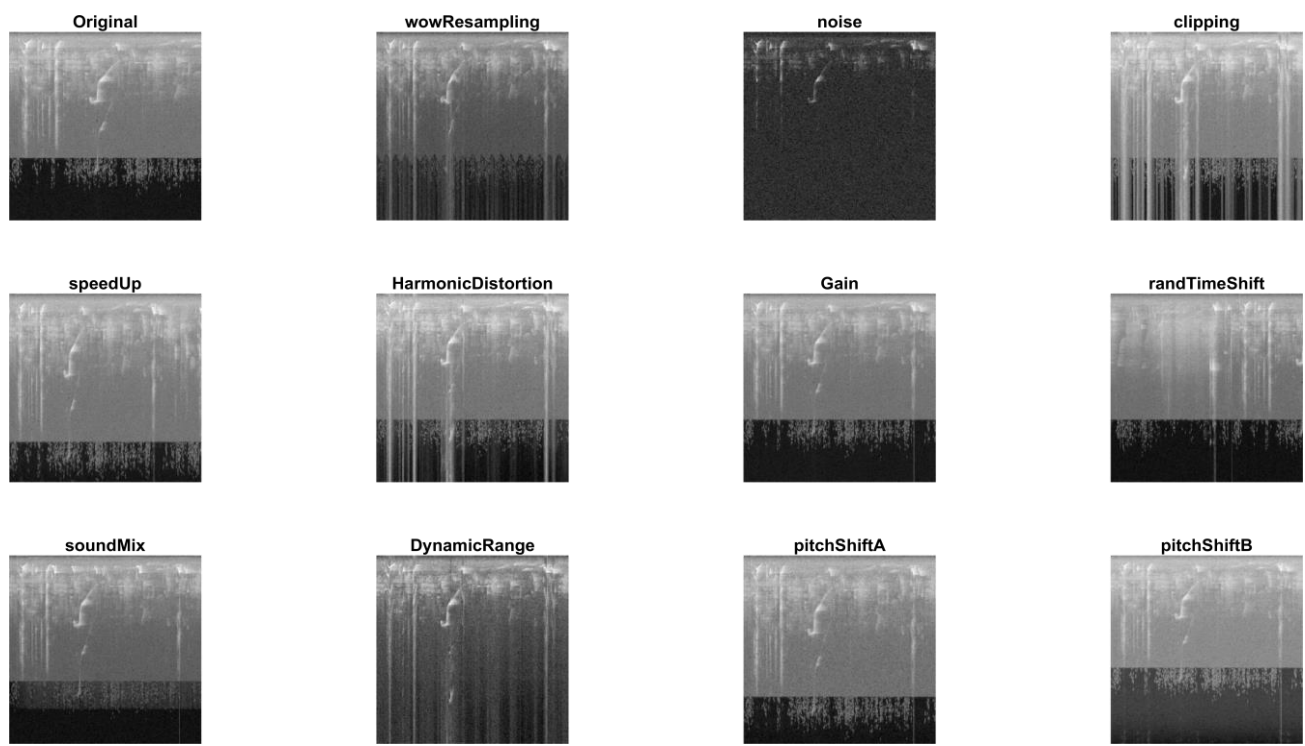

Figure 4. Effect of the 11 transformations included in the Signal augmentation protocol and applied to one illustrative raw audio signal before its conversion into spectrogram.

## 5. Experimental results

We assessed the effects of data augmentation using a stratified ten-fold cross validation protocol and the recognition rate as the performance indicator (i.e. the average accuracy over the different folds). We tested our approach on the following two datasets of animal audio recordings:

- BIRDZ, the control and real-world audio dataset used in [1]. The real-world recordings were downloaded from the Xeno-canto Archive (http://www.xeno-canto.org/), selecting a set of 11 widespread North American bird species. The classes are: 1) Blue Jay, 2) Song Sparrow, 3) Marsh Wren, 4) Common Yellowthroat, 5) Chipping Sparrow, 6) American Yellow Warbler, 7) Great Blue Heron, 8) American Crow, 9) Cedar Waxwing, 10) House Finch and 11) Indigo Bunting. The dataset includes different types of spectrograms: constant frequency, frequency modulated whistles, broadband pulses, broadband with varying frequency components and strong harmonics. Globally, BIRDZ contains 2762 bird acoustic events with 339 detected "unknown" events corresponding to noise and other unknown species vocalizations.

- CAT, the cat sound dataset was presented in [28,50]. It includes 10 balanced sound classes (about 300 samples/class). The classes are: 1) Resting, 2) Warning, 3) Angry, 4) Defence, 5) Fighting, 6) Happy, 7) Hunting mind, 8) Mating, 9) Mother call and 10) Paining. The average duration of a sound is about 4s. The author of this dataset collected the cat sounds from different sources: Kaggle, Youtube and Flickr.

In the following Tables 1 and 2 we report the accuracy obtained by the four data augmentation protocols, comparing them with no augmentation (NoAUG) as baseline, for the CAT and BIRDZ datasets, respectively.

We trained the CNNs for 30 epochs, except for StandardIMG where due to its slow convergence we have run the training for 60 epochs. We used a batch size of 30 for NoAUG and 60 for all the other protocols, to reduce the training time. The learning rate (LR) was set to 0.0001, except for the two GoogleNets in StandardIMG (we used LR=0.001 due to their low performance with LR=0.0001). The CNN named 'VGG16 – batchSize' is the standard VGG16 where the batch size is always fixed to 30.

Moreover, in Tables 1 and 2 we reported also five fusion approaches, based on the assumption that "the collective decision produced by the ensemble is less likely to be in error than the decision made by any of the individual networks" [51]:

1. GoogleGoogle365, sum rule among GoogleNet and GoogleNet – places 365 trained using each of the data augmentation protocols;

2. Fusion – Local, sum rule among the five CNNs trained using each of the data augmentation protocols;
3. Fusion No+Si+Sp, sum rule among fourteen CNNs, i.e. the four CNNs trained with NoAUG, the five trained with Spectro and the five trained with Signal (for each protocol a different training is performed);
4. Fusion Si+Sp, as the previous fusion, but not considering the NoAUG CNNs. Only the five CNNs trained with Spectro and the five trained with Signal are combined by sum rule;
5. Fusion Si+Sp+SSG, as the previous fusion, with the addition of the five CNNs trained with the augmentation protocol StandardSGN.

VGG16 could show a convergence problem: if it did not converge in the training phase, we re-run the training a second time. To avoid numeric problems in the fusions by sum rule, all the scores with not-a-number value were considered as zero. Another numeric problem is that VGG16 can assign the same scores to all the patterns, e.g. when VGG16 does not converge in the training data (random performance in the training set). Also in this case we considered all the scores as zeros.

| CAT | NoAUG | StandardIMG | StandardSGN | Signal | Spectro |
|---|---|---|---|---|---|
| GoogleNet | 82.98 | 76.44 | 85.12 | 85.25 | 88.68 |
| VGG16 | 84.07 | 77.02 | 86.64 | 88.20 | 90.71 |
| VGG19 | 83.05 | 78.47 | 85.59 | 86.71 | 90.68 |
| GoogleNet – places365 | 85.15 | 72.20 | 86.34 | 88.27 | 89.19 |
| VGG16 - batchSize | --- | 78.15 | 84.71 | 88.47 | 91.22 |
| GoogleGoogle365 | 85.86 | 78.15 | 87.83 | 88.34 | 89.83 |
| Fusion - Local | 87.36 | 82.71 | 89.22 | 89.05 | **91.73** |
| Fusion No+Si+Sp | 90.14 | | | | |
| Fusion Si+Sp | 91.08 | | | | |

| | |
|---|---|
| **Fusion Si+Sp+SSG** | 90.71 |

**Table 1.** Performance on the cat dataset (mean accuracy over the ten-fold cross validation). StandardIMG: combination standard image augmentation techniques. StandardSGN: combination of standard MATLAB functions for audio signal processing. Spectro: combination of seven functions working on the spectrograms of the audio signals. Signal: combination of 11 methods for processing of raw audio signals. *convergence problems of VGG16.

| BIRDZ | NoAUG | StandardIMG | StandardSGN | Signal | Spectro |
|---|---|---|---|---|---|
| GoogleNet | 92.41 | 83.76 | 94.66 | 95.32 | 90.51 |
| VGG16 | 95.30 | 91.45 | 95.59 | 95.88 | 90.83 |
| VGG19 | 95.19 | 91.27 | 95.77 | 96.06 | 92.26 |
| GoogleNet – places365 | 92.94 | 85.85 | 94.81 | 95.51 | 92.41 |
| VGG16 - batchSize | --- | 90.85 | 95.84 | 26.98* | 91.24 |
| GoogleGoogle365 | 94.10 | 86.25 | 95.51 | 95.96 | 93.59 |
| Fusion - Local | 95.81 | 92.89 | 96.16 | 96.56 | 94.30 |
| Fusion No+Si+Sp | 96.72 | | | | |
| Fusion Si+Sp | 96.80 | | | | |
| Fusion Si+Sp+SSG | **96.85** | | | | |

**Table 2.** Performance on the BIRDZ dataset (mean accuracy over the ten-fold cross validation).

The following conclusions can be drawn by the reported results:
1. The best performance trade-off performance on the two tested dataset is obtained by "Fusion Si+Sp".

2. There is not a single data augmentation protocol that outperforms all the others in all the tests. Spectro has the best performance in CAT and Signal in BIRDZ. However Signal outperforms NoAUG in both the datasets;

3. The best stand-alone CNN is VGG16 coupled with Signal, although its performance is clearly lower than those obtained by the ensembles;

4. The standard approaches StandardIMG used in computer vision for image augmentation obtain the worst results, also in comparison to NoAUG. Since spectrograms have time and frequency on their axes, standard image augmentation techniques, e.g. reflection, will hardly make sense. For example, if a bird call is characterized by an increasing pitch over the duration of the call, its spectrogram augmented by reflection would represent a completely different sound where the pitch decreases over the duration of the call. This shows the importance of using specific augmentation techniques for audio signals and their spectrograms.

In the following Table 3 we compare our best approach Fusion Si+Sp with our Fusion-Local and the literature data, showing how it outperforms the state of the art performance in both the datasets.

| Descriptor | BIRDZ | CAT |
|---|---|---|
| Fusion - Local Signal | 96.6 | 89.1 |
| Fusion - Local Spectro | 94.3 | **91.7** |
| Fusion Si+Sp | **96.8** | 91.1 |
| [52] | 96.3 | --- |
| [3] | 95.1 | --- |
| [1] | 93.6 | --- |
| [50] | --- | 87.7 |
| [28] | --- | 91.1 |
| [28] - CNN | --- | 90.8 |
| [53] | 96.7* | --- |

**Table 3.** Comparison of Fusion Si+Sp with our Fusion-Local and literature data. *: Inaccurate comparison, the authors used a different testing protocol.

Of note, the comparison with [53] is inaccurate. We used ten-fold cross validation while in that work a different testing protocol was used: in each of 10 trials, they randomly split the dataset into 60% training and 40% test.

We report the results of two approaches extracted from Pandeya et al., named [28] and [28] – CNN, where the latter is based on an ensemble of CNNs for feature extraction to represent the audio signals.

Unfortunately, in the field of audio animal classification, several papers focus only on a single dataset. Here, we have evaluated our protocols on two different datasets to provide a more robust analysis of their general performance. Nevertheless, both datasets tested in this paper were freely available and they were tested here with a clear and unambiguous testing protocol. In this way we report a baseline performance for the audio classification that can be used to compare other methods developed in the future. In terms of computing time, the most expensive activity is the conversion of the audio signal into its spectrogram since it is run on the CPU and not on GPUs. A single audio file of 1.13 seconds required 0.28 seconds for conversion on a machine equipped with i7-7700HQ 2.80 GHz processor, 16 GB RAM and GTX 1080 GPU. The average classification time for a single spectrogram was 0.03 seconds on GoogleNet, 0.01 seconds on VGG16 and 0.01 seconds on VGG19.

**Conclusion**

In this paper we explored how different data augmentation techniques improve the accuracy of automated audio classification of natural sounds (bird and cat sounds) by means of deep network. Different types of data augmentation approaches for audio signals were proposed, tested and compared. Because of the nature of these signals, data augmentation methods were applied on both on the raw audio signals and on their visual representation as spectrogram. A set of CNNs was trained using different sets of data augmentation approaches (that we organized in four protocols), then these CNNs were combined by sum rule.

Our results demonstrated that an ensemble of different fine-tuned CNNs maximizes the performance in the two tested audio classification problems, outperforming previous state-of-the-art approaches. To the best of our knowledge this is the largest study of data augmentation for CNNs in audio classification. We also want to warn about the use of standard data augmentation protocols: augmentation techniques specifically developed for images could be useless, or in the worst case detrimental, when classifying specific dataset. For example, our results show that StandardIMG performs worse than classification with no augmentation. Therefore, when choosing an augmentation protocol for a classification problem, the nature of the dataset to classify must be always taken into consideration.

This work will be further developed by assessing which of the proposed augmentation methods, here grouped in Standard Image Augmentation and Standard Signal Augmentation, are the most beneficial to classify animal sounds, and which ones can be excluded. We also aim i) to include other datasets, e.g. [27,54], in order to obtain a more comprehensive validation of the proposed ensemble of CNNs, ii) to test our ensemble on other sound classification tasks (e.g. whale and frog classification) as well as iii) to assess how different CNN topologies, parameters in the re-tuning step, and data augmentation methods could improve or degrade the ensemble performance.

The MATLAB code for the methods presented in this paper is freely available for comparison at https://github.com/LorisNanni .

**Acknowledgment**

The authors thank NVIDIA Corporation for supporting this work by donating Titan Xp GPU and the Tampere Center for Scientific Computing for generous computational resources.